\documentclass[graybox]{svmult}
\usepackage{mathptmx}       
\usepackage{helvet}         
\usepackage{courier}        
\usepackage{type1cm}        
\usepackage{makeidx}         
\usepackage{graphicx}        
\usepackage{multicol}        
\usepackage[bottom]{footmisc}

\usepackage[utf8]{inputenc}
\usepackage[english]{babel}
\usepackage{amsmath}
\usepackage{algorithm}
\usepackage{algorithmic}
\usepackage{graphicx}
\usepackage{multirow}
\usepackage{url}
\newcommand{\g}{\gamma}
\renewcommand{\t}{\tau}
\newcommand{\s}{\mathit{sup}}
\renewcommand{\c}{\mathit{conf}}
\renewcommand{\D}{\mathcal{D}}
\newcommand{\U}{\mathcal{U}}
\newcommand{\ra}{\rightarrow}
\newcommand{\bs}{\backslash}
\newcommand{\st}{\bigm|}

\newcommand{\mxs}[1]{\mathit{mxs}_{#1}}
\newcommand{\kmns}[1]{\mathit{mns}_{#1}}
\newcommand{\bmns}[1]{\mathit{bmns}_{#1}}
\newcommand{\mxgs}[1]{\mathit{mxgs}_{#1}}
\newcommand{\Fst}[1]{\mathit{F}_{#1}}
\newcommand{\FCst}[1]{\mathit{FC}_{#1}}
\newcommand{\FGst}[1]{\mathit{FG}_{#1}}
\newcommand{\RIst}[1]{\mathit{RI}_{#1}}
\newcommand{\RRst}[1]{\mathit{RR}_{#1}}
\newcommand{\ARst}[1]{\mathit{AR}_{#1}}
\newcommand{\CRIst}[1]{\overline{\mathit{RI}_{#1}}}
\newcommand{\setImplications}{\mathcal{B}}

\newcommand{\closure}[1]{\overline{#1}}
\newcommand{\cl}[1]{\overline{#1}}
\newcommand{\Bstar}[1]{\mathcal{B}^{*}_{#1}}
\newcommand{\constant}[1]{c_{#1}}
\makeindex             

\begin{document}

\title*{Closed-set-based Discovery of Bases of Association Rules}

\author{Jos\'e L Balc\'azar, Diego Garc\'\i{}a-Saiz, Domingo G\'omez-P\'erez and Cristina T\^{i}rn\u{a}uc\u{a}}
\authorrunning{Balc\'azar et al}
\institute{Jos\'e L Balc\'azar \at Departamento de Matem\'aticas, Estad\'\i{}stica y Computaci\'on\\
  Universidad de Cantabria, Santander, Spain, \email{joseluis.blacazar@unican.es}
\and
Diego Garc\'\i{}a-Saiz
\at Departamento de Matem\'aticas, Estad\'\i{}stica y Computaci\'on\\
Universidad de Cantabria, Santander, Spain, \email{diego.garciasuc@alumnos.unican.es}
\and
Domingo G\'omez-P\'erez \at Departamento de Matem\'aticas, Estad\'\i{}stica
y Computaci\'on\\
Universidad de Cantabria, Santander, Spain, \email{domingo.gomez@unican.es}
\and
Cristina T\^{i}rn\u{a}uc\u{a} \at Departamento de Matem\'aticas,
Estad\'\i{}stica y Computaci\'on\\
Universidad de Cantabria, Santander, Spain, \email{cristina.tirnauca@unican.es}
}

\maketitle

\abstract*{
The output of an association rule miner is often huge in
practice. This is why several concise lossless representations have
been proposed, such as the "essential" or "representative" rules. We
revisit the algorithm given by Kryszkiewicz (Int. Symp. Intelligent
Data Analysis 2001, Springer-Verlag LNCS 2189, 350-359) for mining
representative rules. We show that its output is sometimes incomplete,
due to an oversight in its mathematical validation. We propose
alternative complete generators and we extend the approach to an
existing closure-aware basis similar to, and often smaller than, the
representative rules, namely the basis B*.}
\abstract{
The output of an association rule miner is often huge in
practice. This is why several concise lossless representations have
been proposed, such as the ``essential'' or ``representative'' rules. We
revisit the algorithm given by Kryszkiewicz (Int.~Symp.~Intelligent
Data Analysis 2001, Springer-Verlag LNCS 2189, 350--359) for mining
representative rules. We show that its output is sometimes incomplete,
due to an oversight in its mathematical validation. We propose
alternative complete generators and we extend the approach to an
existing closure-aware basis similar to, and often smaller than, the
representative rules, namely the basis $\Bstar{\t,\g}$.}
\section{Introduction}
Association rule mining is among the most popular conceptual tools in
the field of Data Mining. We are interested in the process of discovering
and representing regularities between sets of items in large scale
transactional data. Syntactically, the association rule representation
has the form of an implication, \emph{$X \to Y$}; however, whereas
in Logic such an expression is true if and only if $Y$ holds whenever
$X$ does, an association rule is a partial implication, in the sense
that it is enough if $Y$ holds \emph{most of the times} $X$ does.

To endow association rules with a definite semantics, we need to
make precise how this intuition of ``most of the times'' is formalized.
There are many proposals for this formalization. One of the frequently
used measures of intensity of this kind of partial implication is its
\emph{confidence}: the ratio between the number of transactions in
which $X$ and $Y$ are seen together and the number of transactions
that contain $X$. In most application cases, the search space is
additionally restricted to association rules that meet a minimal
\emph{support} criterion, thus avoiding the generation of rules
from items that appear very seldom together in the dataset (formal
definitions of support and confidence are given in Section~\ref{sec:ARRR}).

Many association rule miners exists, Apriori (see~\cite{AMSTV96})
being one of the most widely discussed and used.
The major problem shared by all mining algorithms is that, in practice,
even for reasonable support and confidence thresholds, the output is
often huge. Therefore, several concise lossless representations of the
whole set of association rules have been proposed.
These representations are
based on different notions of ``redundancy''.
In one of these, a rule is redundant if it
is possible to compute exactly its confidence
and support from other information such as
the confidences and supports of
other \emph{informative} rules (see~\cite{Krys02,Luxe91,HaBeMe08,PTBSL05});
this is a quite demanding property. We settle
for a weaker version proposed in several works;
informally, in that version, a rule
is \emph{redundant} with respect to another one if its confidence
and support are always greater, in \emph{any} dataset.
To avoid this redundancy, exactly one notion has been
identified in several sources, namely
the \emph{representative rules}; and a closure-aware
variant both of the redundancy notion and of the
redundancy-free basis is given in \cite{Balc10c}
(precise definitions and references are given below).

We focus in this paper on the main results of~\cite{Krys01},
where a purportedly faster algorithm to construct
representative rules is given, and show by an example that that algorithm
is not
guaranteed to always output all representative rules, because it is
based on a property that does not hold in general; namely, 
the characterization of the frequent closed sets that admit a
decomposition into representative rules misses some such sets.
We propose an alternative, complete characterization, leading
us to the proposal of a first alternative algorithm that is
guaranteed to output all the representative rules: we pre-compute,
for each closed set, some parameters that depend on the confidence
and support thresholds, and then use the above mentioned new
characterization to generate all representative rules.
Compared to the potentially incomplete algorithm
in~\cite{Krys01}, this algorithm, guaranteed to be complete,
has a main drawback:
in \cite{Krys01}, the internal local parameters only depend
on the support threshold, but in our
algorithm these parameters depend also on confidence.
Therefore, each time a new confidence threshold is introduced
by the user, the algorithm has to redo all computations.
Thus, we provide a second algorithm, composed of two parts:
the first one is a pre-processing phase, dependent only on
support, in which a subdivision of the
interval $(0,1]$ is associated to each closed itemset, and the second
part uses this partition to determine, for a given value of the
confidence threshold, which are those sets that can generate
representative rules.

Then, we extend the process to a similar basis which
profits from the more powerful redundancy notions
available for full-confidence implications to often
obtain smaller bases in many applications.

There are a couple of subtle differences between one of the usual
definitions of association rule (the one we employ) and the one in
\cite{Krys01}. First, we do allow having rules with empty
antecedent (clearly, all of them have confidence equal to
the normalized support of the consequent). Moreover, we do
not require the inequalities to be strict when imposing a given
support and confidence threshold. This is just a small detail that
comes handy when the user is interested in obtaining the set of all
representative rules of confidence 1. However, we have carefully
tuned all our argumentations in such a way that these differences
are not relevant; for instance, we have chosen a counterexample
that invalidates Property 9 of~\cite{Krys01} independently of
which of the two definitions is used.

The article is structured as follows. In Section~\ref{sec:prel} we
introduce the basic notions and notations that will be used throughout
the paper and part of the contents of~\cite{Krys01}; and we show
that the algorithm provided there is not guaranteed to always provide the
whole set of representative rules. In Section~\ref{sec:RR} we define
new parameters and discuss their usefulness in generating the set of
all representative rules, providing also efficient algorithms for
this task. We describe in Section~\ref{sec:closure} a parallel
development for an alternative basis, often smaller than the
representative rules.
Section~\ref{sec:practice} contains a comparison of our
approach with the one in~\cite{Krys01} on some datasets.
Concluding remarks and further research topics
are presented in Section~\ref{sec:concl}.

\section{Preliminaries}\label{sec:prel}
A given set of available items $\U$ is assumed; subsets of it are called itemsets.
We will denote itemsets by capital letters from the end of the alphabet, and use
juxtaposition to denote union, as in $XY$.
The inclusion sign as in $X\subset Y$ denotes proper subset, whereas
improper inclusion is denoted $X\subseteq Y$.
For a given dataset $\D$, consisting of $n$ transactions,
each of which is an itemset labeled with a unique transaction identifier,
we define the \emph{support} $\s(X)$ of an itemset $X$ as the
ratio between the cardinality of the set of transactions that contain $X$
and the total number of transactions $n$.
An itemset $X$ is called \emph{frequent} if its support
is greater than or equal to some user-defined threshold $\t \in (0,1]$.
We denote by $\Fst{\t}=\{X \subseteq \U \st \s(X) \geq \t\}$
the set of all frequent itemsets.

Given a set $X \subseteq \U$, the \emph{closure $\overline{X}$ of $X$}
is the maximal set (with respect to the set inclusion) $Y \subseteq \U$
such that $X \subseteq Y$ and $\s(X)=\s(Y)$.
It is easy to see that $\overline{X}$ is uniquely defined.
We say that a set $X \subseteq \U$ is \emph{closed} if $\overline{X}=X$.

Closure operators are characterized by the three properties of
extensivity: $X\subseteq\overline{X}$;
idempotency $\overline{\overline{X}}=\overline{X}$;
and
monotonicity:
$\overline{X}\subseteq\overline{Y}$ if $X\subseteq Y$.
Moreover,
intersections of closed sets are closed.
The empty set is closed if and only if no item appears in each and every transaction.

A \emph{minimal generator} is a set $X$ for which all proper subsets have
closures different from the closure of $X$
(equivalently, $X$ is a minimal generator
if and only if $\s(Y)>\s(X)$ for all $Y \subset X$).

Also, $\FCst{\t}=\{X \in \Fst{\t} \st \overline{X}=X\}$ represents the
set of all frequent closed sets, and
$\FGst{\t}=\{X \in \Fst{\t} \st \forall Y\subset X, \s(Y)>\s(X)\}$
is the set of all frequent minimal generators. Note that $\FCst{\t}$
constitutes a concise lossless representation of frequent itemsets, since
knowing the support of all sets in $\FCst{\t}$ is enough to retrieve the
support of all sets in $\Fst{\t}$.


\begin{example}\label{ex:first}
Let $\D$ be the dataset  represented in Table \ref{tab:firsttable} where
the universe $\U$ of attributes is $\{a,b,c,d,e,f\}$, and consider the
threshold $\t=0.15$.
Clearly, all subsets of $\U$ are frequent,
$\FCst{\t}$ = $\{\emptyset,a,b,c,ab,ac,ad,bc,abcde,abcdef\}$ and
$\FGst{\t}$ = $\{\emptyset, a, b, c, d,e, f, ab, ac, bc, bd, cd, abc\}$
(we abuse the notation and denote sets  by the juxtaposition of their
constituent elements).

\begin{table}[h]
\caption{{Dataset $\D$}}\label{tab:firsttable}
\begin{tabular}{|cccccc|}
	\hline
	$a$ & $b$ & $c$ &$d$ & $e$ & $f$\\
	\hline
	1&1&1&1&1&1\\
	1&1&1&1&1&0\\
	1&1&0&0&0&0\\
	1&0&1&0&0&0\\
	0&1&1&0&0&0\\
	1&0&0&1&0&0\\	
	\hline
\end{tabular}
\end{table}
\end{example}

\subsection{Association Rules and Representative Rules}\label{sec:ARRR}
Given $X$ in $\Fst{\t}$,
the following two notions
were introduced
in~\cite{Krys01} (with longer names):
\begin{eqnarray*}
  \mxs{\t}(X)&=&\max(\{\s(Z) \mid Z \in \FCst{\t}, Z \supset X\} \cup \{0\}),\\
  \kmns{\t}(X)&=&\min(\{\s(Y) \mid Y \in \FGst{\t}, Y \subset X\} \cup \{ \infty\}).
\end{eqnarray*}

That is, $\mxs{\t}(X)$ represents the maximum support of all proper frequent closed
supersets of $X$, and $\kmns{\t}(X)$ is the minimum support of minimal generators
that are proper subsets of $X$. The extra $0$ and $\infty$ are added in order to
make sure that $\mxs{\t}(X)$ and $\kmns{\t}(X)$ are defined even for the cases in
which $X$ has no proper supersets that are frequent and closed, or when it does not
have proper subsets that are minimal generators.
It is easy to check that $\mxs{\t}(X) \leq \s(X) \leq \kmns{\t}(X)$.
Moreover, in \cite{Krys01} it is shown that:

\begin{proposition}
Given $\t \in (0,1]$ and
an itemset $X\in\Fst{\t}$, $X$ is closed if and only if
$\s(X) > \mxs{\t}(X)$ and $X$ is a minimal generator if and only if
$\s(X) < \kmns{\t}(X)$.
\end{proposition}

The association rules considered in this work are implications of
the form $X\to Y$, where $X,Y \subseteq \U$, $Y \neq \emptyset$ and
$X \cap Y=\emptyset$. In~\cite{Krys01}, rules with $X=\emptyset$ are disallowed,
but we do permit them as in practice such rules often play a useful role related
to coverings, described below.  The \emph{confidence} of
$X\to Y$ is $\c(X\to Y) = \s(XY)/\s(X)$, and its \emph{support}
is $\s(X\to Y) = \s(XY)$. The problem of mining association rules
consists in generating all rules that meet the minimum support and confidence
threshold criteria, i.~e.~enumerate the following set:
$\ARst{\t,\g}=\{X \to Y \st \s(X \to Y) \geq \t, \c(X \to Y) \geq \g\}$.

Since the whole set of association rules is quite big in real-world applications,
a number of formalizations of the notion of \emph{redundancy}
among association rules have been introduced
(see~\cite{AggYu01b,Balc10c,Krys98,PTBSL05,Phan01,Luxe91,Zaki04,CriSim02},
the survey~\cite{Krys02}, and Section 6 of~\cite{CegRod06}).
In one common approach,
the \emph{cover} set $C(X \to Y)$ of a rule $X \to Y$ is defined by
$C(X \to Y)=\{X' \to Y' \st X \subseteq X' \textrm{ and } X'Y' \subseteq XY\}$.
Such rules $X'\to Y'$ are redundant with respect to $X\to Y$
in the following sense
(see~\cite{AggYu01b,Krys98} and
 also \cite{Krys98a,Balc10c,Phan01}):

\begin{proposition}
\label{propcover}
Let $r,r'$ be association rules.
Then $r' \in C(r)$ implies $\s(r') \geq \s(r)$ and $\c(r') \geq \c(r)$.
\end{proposition}

In fact, this implication is a full characterization,
that is, if $r'$ has always at least the same confidence
and at least the same support as $r$ then it must
belong to the cover set. 
Avoiding such redundancies leads to
the set $\RRst{\t,\g}$ of \emph{representative association rules}.
A rule $r$ in $\ARst{\t,\g}$ is said to be \emph{representative},
or \emph{essential},
if it is not contained in the cover set of any other rule in $\ARst{\t,\g}$, i. e.


\begin{equation*}
\RRst{\t,\g}=\{r \in \ARst{\t,\g} \st \forall r' \in \ARst{\t,\g} \, (r \in C(r') \Rightarrow r=r')\}.
\end{equation*}

\begin{proposition}
The following properties hold:
\begin{itemize}
\item $\RRst{\t,\g}=\{X \to Y \in \ARst{\t,\g} \st \neg \exists X' \to Y' \in \ARst{\t,\g}$, $(X=X',XY \subset X'Y')$ or $(X' \subset X,XY=X'Y')\}$
\item if $X \to Z \bs X$ with $X \subset Z$ is in $\RRst{\t,\g}$ then $Z \in \FCst{\t}$  and $X \in \FGst{\t}$.
\end{itemize}
\end{proposition}

Therefore, any algorithm that aims at the discovery of all representative rules should consider only rules of the form $X \to Z \bs X$ with $X \subset Z$, $Z \in \FCst{\t}$ and $X \in \FGst{\t}$. Clearly, not all sets in $\FCst{\t}$ can be decomposed in such a way, and one should look only into those that do.

\begin{example}\label{ex:trulysecond}
Consider 
the dataset 
in Example \ref{ex:first}. 
The set $ad$ is both frequent and closed,
but none of the rules $a \to d$, $d \to a$ or $\emptyset \to ad$ are
representative given the thresholds $\t=0.15$ and $\g=0.33$:
$a \to d$ is in the cover set of $a \to bd$, $d \to a$ is in the cover
set of $d \to ab$ and $\emptyset \to ad$ is in the cover set of $\emptyset \to abd$.
Also, it is easy to check that, at $\t=0.15$ and $\g=0.4$,
one can obtain representative rules exactly out of
the following closed sets: $ab$, $ac$, $ad$, $bc$, $abcde$, and $abcdef$.
\end{example}

So, if we denote by $\RIst{\t,\g}$ the set of all frequent closed itemsets from
which at least one representative rule can be generated, one possible approach to
representative rule mining is to synthesize first the set $\RIst{\t,\g}$, and
then, for each element $Z$ in $\RIst{\t,\g}$, to find
non-empty subsets $X$
such that $X \to Z \bs X$ is representative.
This is precisely the idea behind Algorithm \emph{GenRR} in \cite{Krys01}.
The problem there is that the characterization of the set $\RIst{\t,\g}$ given
by Property 9 of the same paper (on page 355) is incorrect,
possibly leaving out some of the sets that can lead to representative rules.
Namely, it is stated that
$\RIst{\t,\g}=\{X \in \FCst{\t} \st \s(X)\ge \g * \kmns{\t}(X) > \mxs{\t}(X)\}$;
right-to-left inclusion indeed holds, but equality does not hold in general, as
one can see from the following counterexample.

\begin{example}\label{ex:trulythird}
Consider the itemset $X=abcde$ in Example \ref{ex:first}, and
assume $\t=0.15$ and $\g=0.4$. Let us verify that
$abcde \in \RIst{\t,\g} \backslash \{X \in \FCst{\t} \st \s(X)>
\g * \kmns{\t}(X) \geq \mxs{\t}(X)\}$.
Clearly, the rule $b \to acde$ is in $\ARst{\t,\g}$,
having support 2/6 and confidence 0.5.
Moreover, by extending the right-hand side or moving the item $b$
to the right-hand side we get only the rules $b \to acdef$, $\emptyset \to abcde$
and $\emptyset \to abcdef$ of confidence 1/4, 2/6 and 1/6, respectively.
Hence, we can conclude that $b \to acde\in \RRst{\t,\g}$.
On the other hand, $\mxs{\t}(X)=1/6$ and $\kmns{\t}(X)=2/6$,
so $\g * \kmns{\t}(X)=0.8/6$ is strictly smaller than $\mxs{\t}(X)$.
In this case,  Algorithm~\emph{GenRR} does not work correctly since it
does not list the rule $b \to acde$ as being representative.
\end{example}
An alternative counterexample is given in the proof of Lemma~\ref{ex:second} below.

\section{Characterizing Representative Rules}
\label{sec:RR}

The goal of pruning off sets that do not give
representative rules, by keeping only
$\RIst{\t,\g}$, cannot be reached using
the bounds given, as we have seen that this
set comprises all  $X$ in $\FCst{\t}$ with
$\s(X) \geq \g * \kmns{\t}(X) > \mxs{\t}(X)$ but may also include other
frequent closed sets $X$ that do not satisfy the condition
$\g * \kmns{\t}(X) > \mxs{\t}(X)$. We consider two alternatives.

\subsection{Closed Sets Instead of Minimal Generators}\label{sec:ClosedSets}


For closed $X$, $\kmns{\t}(X)$ is almost the same thing
as the minimal support among all proper subsets of $X$, or again
among all proper closed subsets of $X$; all these notions coincide
when $X$ is its own minimal generator, otherwise they only differ
due to the minimal generators of $X$. Therefore it makes sense to
try and exclude the minimal generators of $X$ from consideration.
This way, we get another parameter,

\smallskip

\indent\indent
$\bmns{\t}(X)=\min(\{\s(Y) \mid Y \in \FCst{\t}, Y \subset X\} \cup \{ \infty\})$.

\smallskip

The value of $\bmns{\t}$ is never smaller than $\kmns{\t}$ as we shall shortly see.
Thus, there will be more sets that meet the condition $\g * \bmns{\t}(X) > \mxs{\t}(X)$.

\begin{proposition}\label{prop:bmns}
The following properties hold.
\begin{itemize}
\item $\bmns{\t}(X)=\min(\{\s(Y) \mid Y \in \FGst{\t}, \overline{Y} \subset X\} \cup \{ \infty\})$,
\item $\kmns{\t}(X) \leq \bmns{\t}(X)$,
\item if $X \in \FCst{\t} \cap \FGst{\t}$ then $\kmns{\t}(X) = \bmns{\t}(X)$,
\end{itemize}
\end{proposition}

\noindent{\sl Proof.}
We omit the proof of the first two claims because they are straightforward.
So, let $X$ be a frequent closed set that is also a minimal generator.
If $X=\emptyset$, then $\kmns{\t}(X)=\bmns{\t}(X)=\infty$.
Otherwise, let $Y \in \FGst{\t}$ be such that $Y \subset X$ and $\kmns{\t}(X)=\s(Y)$.
Clearly, $\overline{Y} \in \FCst{\t}$ and $\overline{Y} \subseteq \overline{X}=X$.
Since $X \in \FGst{\t}$ and $Y \subset X$, $\s(Y)>\s(X)$ and hence
$\s(\overline{Y})>\s(X)$, and therefore $\overline{Y} \subset X$.
We get $\s(\overline{Y}) \geq \bmns{\t}(X)$ and $\kmns{\t}(X) \geq \bmns{\t}(X)$.
Combining it with the fact that $\kmns{\t}(X) \leq \bmns{\t}(X)$ always holds,
we conclude that $\kmns{\t}(X) = \bmns{\t}(X)$.
\qed

Unfortunately, the new parameter can still leave out some sets in $\RIst{\t,\g}$.
\begin{lemma}\label{ex:second}
$\RIst{\t,\g}
\not\subseteq
\{X \in \FCst{\t} \st \s(X)> \g * \bmns{\t}(X) \geq \mxs{\t}(X)\}$.
\end{lemma}

\noindent{\sl Proof.}
Let $\U=\{a,b,c\}$ and $\D$ be the dataset containing
the following 13 transactions:
$t_1=\cdots=t_8=abc, t_9=ab, t_{10}=t_{11}=t_{12}=a, t_{13}=b$;
assume $\t=0.07$ and $\g=0.7$.
One can check that, although $ab \in \RIst{\t,\g}$
(since $a \to b \in \RRst{\t,\g}$),
both $\bmns{\t}(ab)=10/13$
and $\kmns{\t}(ab)=10/13$;
but
$\gamma*\kmns{\t}(ab) = \gamma*\bmns{\t}(ab) = 7/13 < 8/13 = \mxs{\t}(ab)$.
\qed


The next construction shows that by using $\bmns{\t}$ instead of $\kmns{\t}$ we can even leave out some sets in $\RIst{\t,\g}$ that would not have been left out otherwise.

\begin{lemma}\label{ex:third}
$
\RIst{\t,\g}
\cap
\{X \in \FCst{\t} \st \s(X)> \g * \kmns{\t}(X) \geq \mxs{\t}(X)\}
\not\subseteq
\{X \in \FCst{\t} \st \s(X)> \g * \bmns{\t}(X) \geq \mxs{\t}(X)\}
$.
\end{lemma}

\noindent{\sl Proof.}
{Let $\U=\{a,b,c,d,e\}$ and $\D$ be a
dataset containing
35 transactions: $t_1=t_2=abcde,t_3=t_4=t_5=abcd,t_6\cdots=t_{20}=a$ and $t_{21}= \cdots t_{35}=b$.
Pick $\t=0.05$ and $\g=0.75$.}
{
Note that $ab \to cd \in \RRst{\t,\g}$,
and therefore $abcd \in \RIst{\t,\g}$.
Now, $\kmns{\t}(abcd)=5/35$, $\bmns{\t}(abcd)=20/35$,
$\s(abcd)=5/35$ and $\mxs{\t}(abcd)=2/35$.
Although $\g * \kmns{\t}(abcd) = 3.5/35 = 0.1$
belongs to the interval $\left[2/35, 5/35 \right)$,
$\g * \bmns{\t}(abcd)=15/35$ does not.}
\qed
%

\subsection{Minimal Generators of Bounded Support}
In order to give a complete characterization for the set $\RIst{\t,\g}$,
let us first introduce the following notation: for a set $X$ in $\FCst{\t}$,
$\mxgs{\t,\g}(X)$ is the maximal support of those minimal generators that are
included in $X$ and are not more frequent than $\s(X) \slash \g$:
\begin{equation*}
  \mxgs{\t,\g}(X)=
  \max(\{\s(Y) \st Y \in \FGst{\t}, Y \subset X, \g *
  \s(Y) \leq \s(X)\} \cup \{0\}).
\end{equation*}
Note that $\mxgs{\t,\g}(X)$ is either 0, or it is greater than or equal to $\s(X)$.
We prove two propositions that explain how we can use this value in order to compute
the set $\RIst{\t,\g}$ and how to find,
given $X \in \RIst{\t,\g}$, a subset $X_0 \subset X$ such that
$X_0 \to X \bs X_0 \in \RRst{\t,\g}$.

\begin{proposition}\label{prop:sets}
The following equality holds.
$$\RIst{\t,\g}=\{X \in \FCst{\t} \st \g * \mxgs{\t,\g}(X) > \mxs{\t}(X)\} .$$
\end{proposition}
\noindent{\sl Proof.}
Let $X$ be an arbitrary set in $\RIst{\t,\g}$, and take $X_0$ in $\FGst{\t}$ such
that
$X_0 \subset X$ and
$X_0 \ra X \bs X_0 \in \RRst{\t,\g}$.

We have, on one hand, $\c(X_0 \ra X \bs X_0) \geq \g$, and on the other hand, the
rule should not be in the cover set of any other rule with confidence greater than
$\g$, i. e.
$\c(X_0 \ra Z \bs X_0) < \g$ for all $Z \in \FCst{\t}$ with $Z \supset X$.

That is, $\s(X) \geq \g * \s(X_0) > \s(Z)$ for all $Z \in \FCst{\t}$ with $Z \supset X$. From the first inequality, we deduce that $X_0$ meets all the conditions in order
to be considered for the computation of $\mxgs{\t,\g}(X)$, and therefore,
$\mxgs{\t,\g}(X) \geq \s(X_0)$. From the second, we get $\g * \s(X_0) > \mxs{\t}(X)$.
We conclude that $\g * \mxgs{\t,\g}(X) > \mxs{\t}(X)$.

Conversely, let $X \in \FCst{\t}$ be such that $\g * \mxgs{\t,\g}(X) > \mxs{\t}(X)$.
It is clear that $\mxgs{\t,\g}(X)$ cannot be 0 (since $\mxs{\t}(X) \geq 0$),
so

 $$\{ Y \in \FGst{\t} \st Y \subset X, \g * \s(Y)\leq \s(X)\}\neq \emptyset.$$

Take $X_0 \in \FGst{\t}$ to be a set of maximal support that belongs to that set.
Therefore, we have $\mxgs{\t,\g}(X)=\s(X_0)$. Since $\s(X_0 \ra X \bs X_0)=\s(X) \geq \t$
and $\c(X_0 \ra X \bs X_0)=\frac{\s(X)}{\s(X_0)} \geq \g$ we deduce
that $X_0 \ra X \bs X_0 \in \ARst{\t,\g}$.
Note that for any $Z \supset X$, $\c(X_0 \ra Z \bs X_0) =
\frac{\s(Z)}{\s(X_0)} \leq \frac{\mxs{\t}(X)}{\s(X_0)} =
\frac{\mxs{\t}(X)}{\mxgs{\t,\g}(X)} < \g$.
Moreover, for any $X'_0 \subset X_0$, $\s(X'_0)>\s(X_0)$
(since $X_0 \in \FGst{\t}$) and $\g * \s(X'_0) > \s(X)$
(due to the choice we have made for $X_0$).
This is why  $\c(X'_0 \ra X \bs X'_0) = \frac{\s(X)}{\s(X'_0)} < \g$.
We conclude that $X_0 \ra X \bs X_0 \in \RRst{\t,\g}$ and $X \in \RIst{\t,\g}$.
\qed

The previous proposition characterizes unequivocally $\RIst{\t,\g}$.
Simple arithmetic suffices to check that
Proposition~\ref{prop:sets} identifies
exactly the closed sets from which
representative rules follow as per Example~\ref{ex:trulysecond}.
However, we also need a practical method for identifying the set of representative rules. To this end, we give necessary and sufficient conditions for a subset of an itemset in $\RIst{\t,\g}$ to be the left-hand side of a representative rule (see Proposition \ref{prop:rules}).

\begin{proposition}\label{prop:rules}
Let $X \in \RIst{\t,\g}$,
 $c_1 = \mxs{\t}(X) \slash \g$, $c_2=\s(X) \slash \g$ and $X_0 \subset X$.
Then $X_0 \to X \bs X_0 \in \RRst{\t,\g}$ if and only if
$c_1<\s(X_0)\leq c_2 < \kmns{\t}(X_0).$
\end{proposition}

\noindent{\sl Proof.}
Consider $X \in \RIst{\t,\g}$ and $X_0 \subset X$.
Clearly, $X_0 \to X \bs X_0 \in \RRst{\t,\g}$ if and only if the rule
$X_0 \to X \bs X_0$ is in $\ARst{\t,\g}$ and does not belong to the cover set
of any other rule in $\ARst{\t,\g}$. That is equivalent to:
$\s(X) \geq \t$, $\frac{\s(X)}{\s(X_0)} \geq \g$, $\frac{\s(X)}{\s(X'_0)} < \g$
for all $X'_0 \subset X$ and $\frac{\s(Z)}{\s(X_0)} < \g$ for all $Z \supset X$
that satisfy $\s(Z) \geq \t$.

Now, it is easy to see that:
\begin{itemize}
\item $\s(X) \geq \t$ always holds because $X \in \FCst{\t}$,
\item $\frac{\s(X)}{\s(X_0)} \geq \g$ $\Leftrightarrow$ $\s(X_0) \leq c_2$,
\item $\forall X'_0 \subset X_0:  \frac{\s(X)}{\s(X'_0)} < \g$
  $\Leftrightarrow$ $\frac{\s(X)}{\kmns{\t}(X_0)} < \g$
  $\Leftrightarrow$ $c_2 < \kmns{\t}(X_0)$,
\item $\forall Z \supset X: \left(Z \in \Fst{\t} \Rightarrow \frac{\s(Z)}{\s(X_0)}
    < \g \right)$ $\Leftrightarrow$ $\frac{\mxs{\t}(X)}{\s(X_0)} < \g$
  $\Leftrightarrow$ $c_1 < \s(X_0)$,
\end{itemize}
which concludes the proof.\qed

The correctness of Algorithm~\ref{alg:first} trivially follows from
Propositions~\ref{prop:sets} and ~\ref{prop:rules}.

\begin{algorithm}[h]
\caption{RR Generator}\label{alg:first}
  \begin{algorithmic}[1]
    \STATE Input: support threshold $\t$, confidence threshold $\g$
    \STATE $\Fst{\t}=\{X \subseteq \U \st \s(X) \geq \t\}$
    \STATE $\FCst{\t}=\{X \in \Fst{\t} \st  \overline{X}=X\}$
    \STATE $\FGst{\t}=\{X \in \Fst{\t} \st  \forall Y \subset X, \s(Y) > \s(X)\}$
    \FORALL {$X \in \FGst{\t}$}    	
    	\STATE $\kmns{\t}(X)=\min(\{\s(Y) \st Y \in \FGst{\t}, Y \subset X\} \cup \{ \infty\})$
    \ENDFOR
    \STATE $\RIst{\t,\g}=\emptyset$
    \FORALL {$X \in \FCst{\t} \bs \{ \emptyset\}$}
    	\STATE $\mxs{\t}(X)=\max(\{\s(Z) \st Z \in \FCst{\t}, Z \supset X\} \cup \{0\})$
    	\STATE $\mxgs{\t,\g}(X)=\max(\{\s(Y) \st Y \in \FGst{\t}, Y \subset X, \g * \s(Y)\leq \s(X)\} \cup \{0\})$
    	\IF {$\g * \mxgs{\t,\g}(X) > \mxs{\t}(X)$}
    		\STATE add $X$ to $\RIst{\t,\g}$
    	\ENDIF
    \ENDFOR
    \FORALL {$X \in \RIst{\t,\g}$}
    		\STATE $c_1=\mxs{\t}(X) \slash \g$
    		\STATE $c_2=\s(X) \slash \g$
    		\STATE Ant = $\{X_0 \in \FGst{\t} \st X_0 \subset X, c_1 < \s(X_0) \leq c_2 < \kmns{\t}(X_0)\}$
    		\FORALL {$X_0 \in $ Ant}
    			\STATE output $X_0 \to X \bs X_0$
    		\ENDFOR
    \ENDFOR
  \end{algorithmic}
\end{algorithm}

\subsection{An Algorithm for Different Confidence Thresholds}

The disadvantage of Algorithm~\ref{alg:first}, compared to the one in~\cite{Krys01},
is that, for a given $X$ in $\FCst{\t}$, $\mxgs{\t,\g}(X)$ depends on the confidence
threshold, and hence it cannot be reused once $\g$ has changed, whereas both
$\mxs{\t}(X)$ and $\kmns{\t}(X)$ can be computed only once for a given value of
$\t$ and then used for different confidence values.
On the other hand, Algorithm~\ref{alg:first} is guaranteed not to lose representative rules,
whereas the one in \cite{Krys01} risks giving incomplete output,
as in our counterexamples above.



Instead of computing $\mxgs{\t,\g}(X)$ for each and every $\g$, one can find the
individual points~of the interval $(0,1]$ where $\mxgs{\t,\g}(X)$ changes its value.
Indeed, given $X$ in $\FCst{\t} \bs \{\emptyset\}$, let $\{Y_1,\ldots,Y_{n[X]}\}$
be the set $\{Y \in \FGst{\t} \st Y \subset X\}$ in descending order of support.
It is easy to see that
\[ \mxgs{\t,\g}(X) =
  \left\{
    \begin{array}{ll}
       \s(Y_1),     & \mbox{if $\g \leq \frac{\s(X)}{\s(Y_1)}$,} \\
       \s(Y_{i+1}), & \mbox{if $\g \in \left( \frac{\s(X)}{\s(Y_{i})},\frac{\s(X)}{\s(Y_{i+1})} \right]$, $i \in \{1,\ldots,n[X]-1\}$,}\\
       0,           & \mbox{otherwise}.
    \end{array}
  \right.
\]

Let us introduce the following notation: for $i \in \{1,\ldots,n[X]\}$, $y_i[X]=\s(Y_i)$ and $p_i[X]=\s(X) \slash \s(Y_i)$. Moreover, $p_0[X]=0$.
Now, each time a new value of the confidence threshold $\g$ is given, one can decide
whether a frequent closed set $X$ is in $\RIst{\t,\g}$ by simply retrieving the
interval $\left( p_i[X],p_{i+1}[X] \right]$ with $i \in \{0,\ldots,n[X]-1\}$
to which $\g$ belongs (recall that in this case $\mxgs{\t,\g}(X)=y_{i+1}[X]$) and
then checking whether the inequality $\g * y_{i+1}[X] > \mxs{\t}(X)$ holds.
Note that if no such $i$ exists (that is, whenever $\g$ has a value strictly greater
than $p_{n[X]}[X]$), $\mxgs{\t,\g}(X)$ takes the value 0, which makes
$\g * \mxgs{\t,\g}(X)$ smaller than or equal to $\mxs{\t}(X)$.

These ideas are implemented in Algorithms~\ref{alg:preproc} and~\ref{alg:second}.

\begin{algorithm}[h]
\caption{RR Generator - preprocessing phase\label{alg:preproc}}
  \begin{algorithmic}[1]
    \STATE Input: support threshold $\t$
    \STATE $\Fst{\t}=\{X \subseteq \U \st \s(X) \geq \t\}$
    \STATE $\FCst{\t}=\{X \in \Fst{\t} \st \overline{X}=X\}$
    \STATE $\FGst{\t}=\{X \in \Fst{\t} \st \forall Y \subset X, \s(Y) > \s(X)\}$
    \FORALL {$X \in \FGst{\t}$}    	
    	\STATE $\kmns{\t}(X)=\min(\{\s(Y) \st Y \in \FGst{\t}, Y \subset X\} \cup \{ \infty\})$
    \ENDFOR
    \FORALL {$X \in \FCst{\t} \bs \{\emptyset\}$}
    	\STATE $\mxs{\t}(X)=\max(\{\s(Z) \st Z \in \FCst{\t}, Z \supset X\} \cup \{0\})$
    	\STATE $n[X]=|\{Y \in \FGst{\t} \st Y \subset X\}|$
    	\STATE let $\{Y_1,\ldots,Y_{n[X]}\}$ be the set $\{Y \in \FGst{\t} \st Y \subset X\}$ in descending order of support
    	\FORALL {$i \in \{1,\ldots, n[X]\}$}
    		\STATE $y_i[X]=\s(Y_i)$
    		\STATE $p_i[X]=\s(X) \slash y_i[X]$
    	\ENDFOR
    	\STATE $p_0[X]=0$
    \ENDFOR
  \end{algorithmic}
\end{algorithm}

\begin{algorithm}[h]
\caption{RR Generator - second phase\label{alg:second}}
  \begin{algorithmic}[1]
    \STATE Input: support threshold $\t$, confidence threshold $\g$
    \STATE $\RIst{\t,\g}=\emptyset$
    \FORALL {$X \in \FCst{\t} \bs \{\emptyset\}$}
    	\IF {$\exists i \in \{0,\ldots,n[X]-1\}$ such that $\g \in \left( p_i[X],p_{i+1}[X] \right]$}
   			\IF {$\g * y_{i+1}[X]  > \mxs{\t}(X)$ }
   				\STATE add $X$ to $\RIst{\t,\g}$
   			\ENDIF
   		\ENDIF
    \ENDFOR
    \FORALL {$X \in \RIst{\t,\g}$}
    		\STATE $c_1=\mxs{\t}(X) \slash \g$
    		\STATE $c_2=\s(X) \slash \g$
    		\STATE Ant = $\{X_0 \in \FGst{\t} \st X_0 \subset X, c_1 < \s(X_0) \leq c_2 < \kmns{\t}(X_0)\}$
    		\FORALL {$X_0 \in $ Ant}
    			\STATE output $X_0 \to X \bs X_0$
    		\ENDFOR
    \ENDFOR
  \end{algorithmic}
\end{algorithm}

\section{Characterizing the Basis for Closure-Based Redundancy}
\label{sec:closure}
The results of the previous sections can be extended to find a list of rules
such that any other rule in $\ARst{\t,\g}$ is redundant with respect to one
rule in our list and the set of full-confidence implications. This is
exactly the idea behind  a basis for closure-based redundancy
\cite{Balc10c}.

Let $\setImplications$ be a set of implications, i. e. rules
  that hold with confidence 1.
Partial rule $X' \to Y'$ is closure-based redundant
    relative to $\setImplications$ with respect to $X \to Y$
if any dataset $\D$
    in which all the rules in $\setImplications$ hold with confidence 1
    gives $\c(X'\to Y')\ge \c(X\to Y)$.

Closure-based redundancy and standard redundancy coincide when the set of implications
$\setImplications$ is empty. Knowing the set $\setImplications$
is equivalent to knowing how the closure operator
works on each set.
If the set of implications is empty,
then any subset is closed and all the closure-related
argumentations trivialize; in particular,
in this case
the set of representative rules forms a minimum-size basis.

In any case, we have the following characterization for closure-based redundancy:
\begin{theorem}[\cite{Balc10c}]
  \label{thm:redundancy}
  Let $\setImplications$ be a set of exact rules, with associated closure operator
  mapping each itemset $Z$ to its closure $\closure{Z}$. Let $X'\to Y'$ be a rule
  not implied by $\setImplications$, that is, $Y'\not\subset \closure{X'}$, then
  the following are equivalent:
  \begin{enumerate}
  \item $X\subseteq \closure{X'}$ and $X'Y'\subseteq \closure{X Y}$,
  \item
The rule $X' \to Y'$ is closure-based redundant
    relative to $\setImplications$ with respect to $X \to Y$.
  \end{enumerate}
\end{theorem}

Note that $Y'\not\subset \closure{X'}$ is equivalent to saying that $X'\to Y'$ is not a full implication.
One can then analogously define the closure-based cover set of a rule $X \to Y$ by $\closure{C}(X \to Y)=\{X' \to Y' \st X \subseteq \closure{X'} \textrm{ and } X'Y' \subseteq \closure{XY}\}$. 
Accordingly, we must refine the notion of ``different'' rule since
only the closures are relevant: 
A rule $X' \to Y'$ is closure-equivalent
(again relative to $\setImplications$) to $X \to Y$
when $\closure{X'} = \closure{X}$ and $\closure{X'Y'} = \closure{XY}$.

The minimum-size basis $\Bstar{\t,\g}$ for closure-based redundancy contains all rules in $\ARst{\t,\g}$ of confidence strictly smaller than 1 that are not closure-based redundant with respect to any rule in $\ARst{\t,\g}$, unless 
they are closure-equivalent (see \cite{Balc10c} for details).
Again the main property of this basis is that every rule 
in $\ARst{\t,\g}$ is closure-based redundant with a rule in the basis.

\begin{proposition}\label{prop}
If a rule is not in the basis, then it is closure-based redundant with respect to a rule in the basis that is not closure-equivalent to it.
\end{proposition}

\noindent{\sl Proof.}
Indeed, if $X\to Y\bs X$ is not in the basis, some rule 
$X'\to Y'\bs X'$ exists above
the confidence and support thresholds for which 
$X'\subseteq \closure{X}$ and $Y\subseteq \closure{Y'}$, and either
$\closure{X'} \neq \closure{X}$ or $\closure{Y'} \neq \closure{Y}$;
in turn, this rule is closure-based redundant with a rule in the basis, 
possibly itself, say $X''\to Y''\bs X''$, so that 
$X''\subseteq \closure{X'} \subseteq \closure{\closure{X}} = \closure{X}$ 
and $Y\subseteq \closure{Y'} \subseteq \closure{\closure{Y''}} = \closure{Y''}$;
further, then,
$\closure{X''} = \closure{X}$ implies $\closure{X'} = \closure{X}$,
and $\closure{Y''} = \closure{Y}$ implies $\closure{Y'} \neq \closure{Y}$.
Therefore, if $X\to Y\bs X$ is not in the basis, then it 
is closure-based redundant with $X''\to Y''\bs X''$, 
which is in the basis and is not closure-equivalent to it.
\qed

It is easy to check that, in all rules in this basis,
the left-hand sides are also closed sets. 
We are interested in computing this basis fast.
To do that, let $\CRIst{\t,\g}$ be the
set of all frequent closed itemsets from which
at least one rule for this basis can be obtained.


\begin{proposition}
The following equality holds.
  \begin{equation*}
    \CRIst{\t,\g}=\{X\in\FCst{\t}\ |\ \g * \mxgs{\t,\g}(X) > \mxs{\t,\g}(X)
    \text{ and } \mxgs{\t,\g}(X)>\s(X)\}.
  \end{equation*}
\end{proposition}

\noindent{\sl Proof.} 
Let $X$ be an arbitrary set in $\CRIst{\t,\g}$: there is
a basis rule $X_0 \ra X \bs X_0$ for these confidence and
support thresholds, where $X_0$ is a proper closed
subset $X_0 \subset X$. Pick a minimal generator $X_1$ of
$X_0$; as $X_0$ is closed, $\s(X_1) = \s(X_0) > \s(X)$;
as $\c(X_0 \ra X \bs X_0)\geq\gamma$,
$\g * \s(X_1) = \g * \s(X_0) \leq \s(X)$, hence
$X_1$ participates in the computation of
$\mxgs{\t,\g}(X)$, so that
$\mxgs{\t,\g}(X)\geq\s(X_1)>\s(X)$.

Besides, if there was a proper closed superset $Z$ of $X$ such that
$\s(Z)\geq\tau$
and $c(X_0 \ra Z \bs X_0)\geq\gamma$, then the
rule $X_0 \ra X \bs X_0$ would not be in the basis due to
redundancy with $X_0 \ra Z \bs X_0$. Therefore, the support
of any frequent itemset $Z$ with
$X\subset Z$
is less than $\gamma*\s(X_0)$. That is, $\mxs{\t,\g}(X)<\g * \s(X_0)$. Hence, 
$\g * \mxgs{\t,\g}(X) \geq \gamma*\s(X_1) = \gamma*\s(X_0) >
\mxs{\t,\g}(X)$.

Conversely, assume that
   \begin{equation*}
    \g * \mxgs{\t,\g}(X) > \mxs{\t,\g}(X)\text{ and } \mxgs{\t,\g}(X)>\s(X)
   \end{equation*}
holds for $X\in\FCst{\t}$. Indeed, $\s(X)<\mxgs{\t,\g}(X)$ implies that this last value is not zero, and that there is at least one itemset $X_1 \in \FGst{\t}$ such that $X_1 \subset X$ and $ \g *  \s(X_1) \leq \s(X)$.
Among these $X_1$, we pick one with maximum support:
$\mxgs{\t,\g}(X) = \s(X_1)$. Let $X_0=\cl{X_1}$, so $\s(X_0) = \s(X_1) > \s(X)$ and $X_0 \subset X$.
Then
$\c(X_0 \ra X \bs X_0) = \s(X)/\s(X_0)
 \geq \g * \s(X_1)/\s(X_0) = \g$,
which implies 
$X_0\to X\bs X_0\in\ARst{\t,\g}$.

Suppose, for a contradiction, that $X_0\to X\bs X_0$
is not in the basis. By Proposition~\ref{prop}, it must
be closure-based redundant
with respect
to a rule $Y \to Z\bs Y$ that is in the basis and is not
closure-equivalent to it. Being in the basis implies that
$Y,\ Z\in\FCst{\t}$ (and keep in mind that both
$X_0$ and $X$ are closed as well).
By Theorem~\ref{thm:redundancy}, we have that
$ Y\subseteq X_0$ and $X\subseteq Z$,
where one of the two inclusions must be proper
to ensure closure-inequivalence.
If $X\subset Z$, we have that
$$
    \c(Y \to Z\bs Y)=\frac{\s(Z)}{\s(Y)}\le \frac{\s(Z)}{\s(X_0)}\le
    \frac{\mxs{\t}(X)}{\mxgs{\t,\g}(X)}< \g,
$$
which is a contradiction with $\c(Y \to Z\bs Y)\ge \g$
as $Y \to Z\bs Y \in \Bstar{\t,\g}\subseteq\ARst{\t,\g}$.
The other possibility is that
$Z=X$ and $Y\subset X_0$, but 
$\s(Y)>\s(X_0)$, because $Y\in\FCst{\t}$, 
contradicting the maximality of $\s(X_0)$. 
This finishes the proof.\qed

\begin{proposition}
  Let $X\in\CRIst{\t,\g}$, $\constant{1}=\mxs{\t}(X)/\g$,
and  $\constant{2}=\s(X)/\g$. Consider a proper closed subset 
$X_0 \subset X$. Then $X_0\to X\bs X_0 \in \Bstar{\g}$
  if and only if $\constant{1}<\sup(X_0)\le \constant{2}<\kmns{\t}(X_0)$.
\end{proposition}
\noindent{\sl Proof.} 
Consider $X\in\CRIst{\t,\g}$ and a proper closed subset
$X_0\subset X$. The rule
$X_0\to X\bs X_0$ is in $\Bstar{\g}$ if and only if it meets the support and confidence threshold requirements with respect to $\t$ and $\g$, it is not a full implication, and is not closure-based redundant with
respect to another rule $Y\to Z\bs Y$.
\smallskip
\newline
First of all $\s(X)\ge \t$, because $X\in\CRIst{\t,\g}$ so it remains to  see that:
\begin{enumerate}
\item $\c(X_0\to X\bs X_0)\ge \g$,
\item $\c(Y\to Z\bs Y)< \g$ for any $Y,Z\in\FCst{\t}$ such that
  $Y\subseteq X_0$
  and $X\subseteq Z$, with at least one
  of the two inclusions proper.
\end{enumerate}
The first item is equivalent to $\s(X_0)\le\constant{2}$; for the second item
we will divide the proof in two different steps:
first, we are going to consider the case where $Y\subset X_0$ and
$X\subseteq Z$.
$$  \forall Y\subset X_0,\ \c(Y\to Z\bs Y)<\g \iff \frac{\s(X)}{\s(Y)}<\g
\iff \constant{2}<\kmns{\t}(X_0).
$$
In a similar way, we obtain that for all $Z$ such that $X\subset Z$
and $Y=X_0$, $\c(Y\to Z\bs Y)<\g$ is equivalent to $\constant{1}<\s(X_0)$.
This finishes the proof.\qed

All the three algorithms defined so far can be modified to output the set $\Bstar{\t,\g}$ of closure-based irredundant partial rules.
 These
modifications are easy  from the results we have proven in this Section, so
they are omitted.

\section{Empirical Comparison}\label{sec:practice}
We have seen that one can find toy examples of datasets in which the output of
the algorithm in~\cite{Krys01} is incomplete.

We have tested our algorithms on two real-world datasets:
the training set part of the UCI Adult US census dataset (see \cite{AsuNew07}) and a Retail dataset (see~\cite{brijs99using}).

We have implemented three different algorithms:
one for the incomplete heuristic given in \cite{Krys01},
one that generates the complete set of representative rules
as described by Algorithm~\ref{alg:first}, and the last algorithm
outputs a complete basis under the notion of closure-based redundancy. 
In order to get comparable results, all algorithms allow rules with
empty antecedent and use the same definition of frequent sets and association rules
as given in our preliminaries.
We
emphasize that, in general, the incomplete heuristic fails to produce a complete
basis of representative rules. The code is available at \cite{slatt023}.

The first dataset under study, which we refer by the name of Retail, is a market basket data which
consists of 88163 transactions over 16470 attributes.
In order to preserve the anonymity of the clients, the data has been processed so that each item is represented by a number and each
line break  separates different customers. For the
interested reader, the paper~\cite{brijs99using} contains more information
about this dataset.

Table~\ref{tab:cesta} shows the number of representative rules obtained for
different support and confidence thresholds (the seventh column), 
the  cardinality of the output set when $\kmns{\t}$ is used (the fifth column)
and the time elapsed in order to obtain them (the sixth and forth columns, respectively). We can see that although for
higher support thresholds the output of the algorithms is, most of the times,
identical (recall that the output of the algorithm in~\cite{Krys01} is always
a subset of the whole set of representative rules), lowering both thresholds
shows bigger differences.
\begin{table}[h]
\caption{Comparison between \emph{GenRR} and Algorithm~\ref{alg:first} on the Retail dataset}\label{tab:cesta}

  \begin{tabular}{|c|c|c|c|c|c|c|}
    \hline
    \multicolumn{3}{|c|}{Data}&\multicolumn{2}{|c|}{\emph{GenRR}}&\multicolumn{2}{c|}{Algorithm~\ref{alg:first}}\\\hline
    $|\FCst{\t}|$ &Support&Confidence&Time &Rules &Time &Rules\\\hline
    &&0.9&0.015&248&0.013&248\\\cline{3-7}
    7573&0.1\%&0.8&0.013&643&0.013&652\\\cline{3-7}
    &&0.7&0.028&1978&0.026&1990\\\hline
    &&0.9&0.036&670&0.022&670\\\cline{3-7}
    19115&0.05\%&0.8&0.073&2228&0.041&2229\\\cline{3-7}
    &&0.7&0.123&6029&0.083&6039\\\hline
  \end{tabular}
\end{table}

Dataset Adult is a transactional version of the training set part of the
UCI census dataset Adult US (see \cite{AsuNew07});
it consists of 32561 transactions over 269 items.
On the Adult dataset, we see the same trend in the behavior
of both algorithms.
Note that in this case there are significant differences between the output
of the algorithm in~\cite{Krys01} and the set of all representative rules
(Table~\ref{tab:Adultrain}). For example, for support and confidence
thresholds of 0.05 and 0.7, respectively, more than half of the rules are lost.
\begin{table*}
\caption{{Comparison between \emph{GenRR} and Algorithm \ref{alg:first} on the Adult dataset}}\label{tab:Adultrain}
  \begin{tabular}{|c|c|c|c|c|c|c|}\hline
    \multicolumn{3}{|c|}{Data}&\multicolumn{2}{|c|}{\emph{GenRR}}&\multicolumn{2}{c|}{Algorithm~\ref{alg:first}}\\\hline
    $|\FCst{\t}|$ &Support&Confidence&Time &Rules &Time &Rules\\\hline
    &&0.9&0.147&6578&0.176&7436\\\cline{3-7}
    11920&1\%&0.8&0.130&4827&0.148&7379\\\cline{3-7}
    &&0.7&0.096&3459&0.141&6867\\\hline
    &&0.9&0.391&15208&0.380&17573\\\cline{3-7}
    27444&0.5\%&0.8&0.298&11516&0.417&18190\\\cline{3-7}
    &&0.7&0.263&8241&0.382&16779\\\hline
  \end{tabular}
\end{table*}

As an example, in the case the thresholds for support and confidence are $1\%$ and 0.70, respectively,
there are a total of 6867 representative rules,
among which 3408 are lost when using $\kmns{}$ or $\bmns{}$ 
(four of them listed in bold, the rest of the rules are given as an example):

\smallskip

\begingroup\small\parindent0pt

\textbf{[c:0.75, s:1.03] Private White age: 41 $\Rightarrow$ Male,}

\textbf{[c:0.82, s:2.21] Never-married Unmarried $\Rightarrow$ \ $<$=50K USA},

\textbf{[c:0.70, s:1.47] $<$=50K Assoc-acdm White $\Rightarrow$ Private},

\textbf{[c:0.75, s:3.74] Own-child Private hours-per-week: 40 $\Rightarrow$ \ $<$=50K Never-married USA},

{[c:0.75, s:3.74] Never-married Own-child USA hours-per-week: 40 $\Rightarrow$ \ $<$=50K Private},


{[c:0.87, s:1.03 ] Male Private age: 41 $\Rightarrow$ White}

{[c:0.75, s:1.03 ] Private White age: 41 $\Rightarrow$ Male}

{[c:0.86, s:7.07 ] Exec-managerial Private $\Rightarrow$ USA White}

{[c:0.73, s:1.04 ] Craft-repair Divorced $\Rightarrow$ Male USA White}

{[c:0.75, s:1.68] Not-in-family hours-per-week: 50 $\Rightarrow$ \ $<$=50K}

\endgroup 
\smallskip
As mentioned in the beginning of this section, we have run experiments
in order to see the performance of our algorithm that finds a basis under
closed-based redundancy conditions. The results are in Tables~\ref{tab:retail*} and
\ref{tab:adult*}. Notice that in this case the times are significantly lower.
\begin{table*}
  \caption{Algorithm for Basis $\Bstar{\t,\g}$ (Retail dataset)}\label{tab:retail*}
  \begin{tabular}{|c|c|c|c|}
    \hline
    Support& Confidence & Time & Rules \\ \hline	
    &0.9  &0.006 &233\\ \cline{2-4}
    0.1\%& 0.8&0.007&643\\ \cline{2-4}
    &0.7&0.013&1984\\ \hline	
    &0.9  &0.029&549\\\cline{2-4}
    0.05\%& 0.8&0.024 &2139\\ \cline{2-4}
    &0.7&0.044&6039\\ \hline
  \end{tabular}

\end{table*}
\begin{table*}
  \caption{Algorithm for Basis $\Bstar{\t,\g}$ (Adult dataset)}\label{tab:adult*}
  \begin{tabular}{|c|c|c|c|}
    \hline
    Support& Confidence & Time & Rules\\ \hline	
    &0.9  &0.093 &7103\\ \cline{2-4}
    1\%&0.8  &0.086 &7205\\ \cline{2-4}
    &0.7  &0.082 &6662\\ \hline

    &0.9  &0.243 &16457\\ \cline{2-4}
    0.5\%& 0.8&0.250&17531\\ \cline{2-4}
    &0.7&0.233&16085\\ \hline
  \end{tabular}

\end{table*}

We have run the experiments on an Intel Core i3-330M @ 2,13GHz machine
with 4 GB of RAM running under Microsoft Windows 7 Professional (64 bits).
The running time of all  algorithms were between 6 and 123 milliseconds in
the case of the Retail dataset and between 82 and 417 milliseconds 
for the Adult dataset.
Algorithm~\ref{alg:first}  correctly outputs all representative rules 
at the cost of being  sometimes slower than
the possibly incomplete algorithm of Kryszkiewicz but, in our tests,
the difference was rather irrelevant since the time needed to print 
the results on screen
(a device slower than the CPU) still dominates the process.


It must be noted that the quantity of
representative rules may decrease at
lower confidence or support thresholds.
This phenomenon has been observed and
explained before (see \cite{Balc10c})
and is caused by powerful rules of
a given confidence, say 0.8, that are
filtered out at higher thresholds,
leaving therefore many other rules
as representative, but that force
all of these out of the representative
rules set as they become redundant when the
confidence threshold gets below
0.8 and lets the powerful rule in.


\section{Conclusions}\label{sec:concl}



We have proposed an alternative (complete) solution for the generation of the set
of all representative rules defined in \cite{Krys98} (see Algorithm \ref{alg:first});
we have also shown that the original algorithm was incomplete.
Our approach, which seems to requiere more operations than the one in \cite{Krys01},
has the advantage of being guaranteed to output the whole set of representative rules.

On the other hand, one of its main drawbacks is that we cannot reuse 
the pre-computed values
of the parameters once the user changes the confidence threshold.
Our proposal for fixing this problem involves dividing the process
into two phases (see Algorithm \ref{alg:preproc} and Algorithm \ref{alg:second}).
As a conclusion, depending on whether one is interested in getting complete results
or getting them faster, it is more convenient to use Algorithm \ref{alg:first} 
or the algorithm in \cite{Krys01}.

We have also extended our approach to the similar but different basis
corresponding to closure-based redundancy. Tests were performed in other to confirm
that the algorithm is significantly faster than the previous two.




\bibliographystyle{apalike}
\bibliography{bibfile}

\begin{thebibliography}{}

\bibitem[Aggarwal and Yu, 2001]{AggYu01b}
Aggarwal, C.~C. and Yu, P.~S. (2001).
\newblock A new approach to online generation of association rules.
\newblock {\em IEEE Transactions on Knowledge and Data Engineering},
  13(4):527--540.

\bibitem[Agrawal et~al., 1996]{AMSTV96}
Agrawal, R., Mannila, H., Srikant, R., Toivonen, H., and Verkamo, A.~I. (1996).
\newblock Fast discovery of association rules.
\newblock In {\em Advances in Knowledge Discovery and Data Mining}, pages
  307--328. AAAI/MIT Press.

\bibitem[Asuncion and Newman, 2007]{AsuNew07}
Asuncion, A. and Newman, D. (2007).
\newblock {UCI} machine learning repository.

\bibitem[Balc{\'a}zar, 2010a]{Balc10c}
Balc{\'a}zar, J.~L. (2010a).
\newblock Redundancy, deduction schemes, and minimum-size bases for association
  rules.
\newblock {\em Logical Methods in Computer Science}, 6(2:3):1--33.

\bibitem[Balc{\'a}zar, 2010b]{slatt023}
Balc{\'a}zar, J.~L. (2010b).
\newblock Slatt023.
\newblock \url{https://code.google.com/p/slatt/downloads/list}.

\bibitem[Brijs et~al., 1999]{brijs99using}
Brijs, T., Swinnen, G., Vanhoof, K., and Wets, G. (1999).
\newblock Using association rules for product assortment decisions: A case
  study.
\newblock In {\em Knowledge Discovery and Data Mining}, pages 254--260.

\bibitem[Ceglar and Roddick, 2006]{CegRod06}
Ceglar, A. and Roddick, J.~F. (2006).
\newblock Association mining.
\newblock {\em ACM Comput. Surv.}, 38(2).

\bibitem[Cristofor and Simovici, 2002]{CriSim02}
Cristofor, L. and Simovici, D.~A. (2002).
\newblock Generating an informative cover for association rules.
\newblock In {\em Proc. of the 2002 IEEE International Conference on Data
  Mining (ICDM)}, pages 597--600. IEEE Computer Society.

\bibitem[Hamrouni et~al., 2008]{HaBeMe08}
Hamrouni, T., {Ben Yahia}, S., and {Mephu Nguifo}, E. (2008).
\newblock Succinct minimal generators: Theoretical foundations and
  applications.
\newblock {\em Int. J. Found. Comput. Sci.}, 19(2):271--296.

\bibitem[Kryszkiewicz, 1998a]{Krys98a}
Kryszkiewicz, M. (1998a).
\newblock Fast discovery of representative association rules.
\newblock In Polkowski, L. and Skowron, A., editors, {\em Proc. of the 1st
  International Conference on Rough Sets and Current Trends in Computing
  (RSCTC)}, volume 1424 of {\em Lecture Notes in Artificial Intelligence},
  pages 214--221. Springer-Verlag.

\bibitem[Kryszkiewicz, 1998b]{Krys98}
Kryszkiewicz, M. (1998b).
\newblock Representative association rules.
\newblock In Wu, X., Ramamohanarao, K., and Korb, K.~B., editors, {\em Proc. of
  the 2nd Pacific-Asia Conference on Knowledge Discovery and Data Mining
  (PAKDD)}, volume 1394 of {\em Lecture Notes in Artificial Intelligence},
  pages 198--209. Springer-Verlag.

\bibitem[Kryszkiewicz, 2001]{Krys01}
Kryszkiewicz, M. (2001).
\newblock Closed set based discovery of representative association rules.
\newblock In Hoffmann, F., Hand, D.~J., Adams, N.~M., Fisher, D.~H., and
  Guimar{\~a}es, G., editors, {\em Proc. of the 4th International Symposium on
  Intelligent Data Analysis (IDA)}, volume 2189 of {\em Lecture Notes in
  Computer Science}, pages 350--359. Springer-Verlag.

\bibitem[Kryszkiewicz, 2002]{Krys02}
Kryszkiewicz, M. (2002).
\newblock Concise representations of association rules.
\newblock In Hand, D.~J., Adams, N.~M., and Bolton, R.~J., editors, {\em Proc.
  of the ESF Exploratory Workshop on Pattern Detection and Discovery}, volume
  2447 of {\em Lecture Notes in Computer Science}, pages 92--109.
  Springer-Verlag.

\bibitem[Luxenburger, 1991]{Luxe91}
Luxenburger, M. (1991).
\newblock Implications partielles dans un contexte.
\newblock {\em Math\'{e}matiques et Sciences Humaines}, 29:35--55.

\bibitem[Pasquier et~al., 2005]{PTBSL05}
Pasquier, N., Taouil, R., Bastide, Y., Stumme, G., and Lakhal, L. (2005).
\newblock Generating a condensed representation for association rules.
\newblock {\em J. Intell. Inf. Syst.}, 24(1):29--60.

\bibitem[Phan-Luong, 2001]{Phan01}
Phan-Luong, V. (2001).
\newblock The representative basis for association rules.
\newblock In Cercone, N., Lin, T.~Y., and Wu, X., editors, {\em ICDM}, pages
  639--640. IEEE Computer Society.

\bibitem[Zaki, 2004]{Zaki04}
Zaki, M.~J. (2004).
\newblock Mining non-redundant association rules.
\newblock {\em Data Min. Knowl. Discov.}, 9(3):223--248.

\end{thebibliography}
\end{document}